\documentclass{article}
\usepackage{spconf,amsmath,amssymb,graphicx,hyperref}
\usepackage[normalem]{ulem}
\usepackage{array}
\usepackage{multirow}
\usepackage{booktabs}
\usepackage[table]{xcolor}
\usepackage{url}

\title{SGP-SAM: Self-Gated Prompting for Transferring 3D Segment Anything Models to Lesion Segmentation}
\name{Zequan Yao$^{1}$$^{\ast}$ \qquad Zixuan Tang$^{1}$\sthanks{These authors contributed equally.} \qquad Jielin Ma$^{1}$ \qquad Jianjia Zhang$^{2}$ \qquad Shen Zhao$^{1}$\sthanks{Corresponding author: zhaosh35@mail.sysu.edu.cn}}
\address{$^{1}$School of Intelligent Systems Engineering, Sun Yat-sen University, Shenzhen, China \\
$^{2}$School of Biomedical Engineering, Sun Yat-sen University, ShenZhen, China}

\begin{document}
\ninept
\maketitle

\begin{abstract}
Three-dimensional lesion segmentation with promptable foundation models is challenging because small, irregular lesions are poorly represented and foreground voxels are highly imbalanced. We propose SGP-SAM, a feature-gated adaptation framework for transferring SAM-Med3D to volumetric lesion segmentation. The Self-Gated Prompting Module (SGPM) derives a scalar gate from internal features and controls a multi-scale fusion residual path at the module level. We further introduce Zoom Loss, which combines Dice supervision with lesion-size-aware, voxel-balanced focal weighting to emphasize small lesions. Across the two MSD tasks, SGP-SAM consistently improves over SAM-Med3D pretraining, adapter tuning, and full fine-tuning, reaching 0.7151 mDice on MSD Liver Tumor with a 7.3\% relative gain over full fine-tuning.
\end{abstract}

\section{Introduction}
\begin{figure}[t]
\centering
\includegraphics[width=\linewidth]{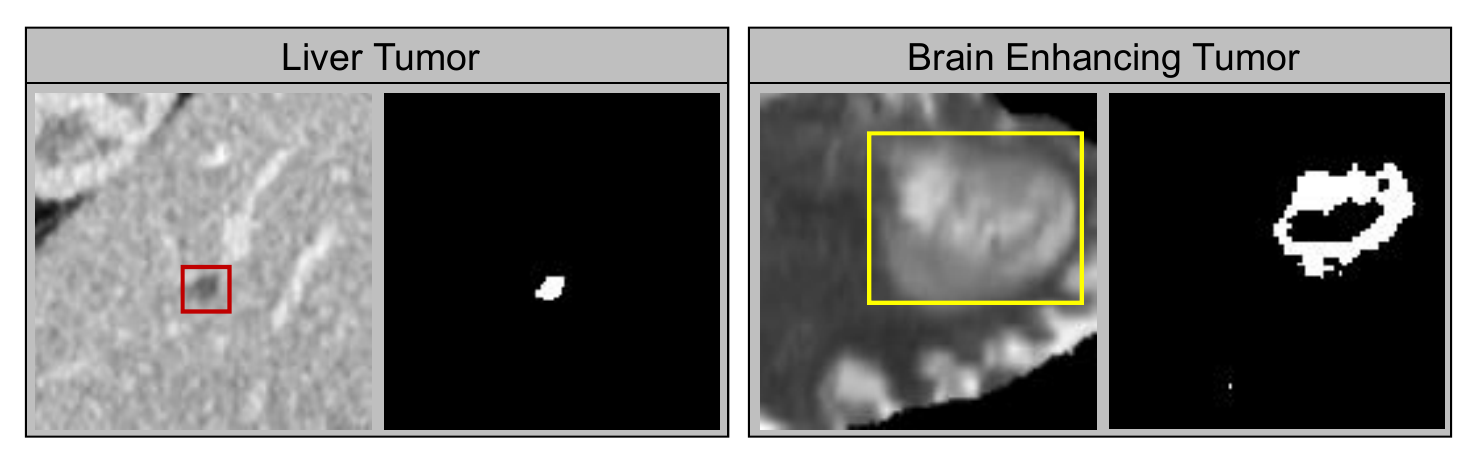}
\caption{Challenges in lesion segmentation: liver tumors occupy a small fraction of CT volumes, while enhancing brain tumors exhibit complex shapes.}
\label{fig:1} 
\end{figure}

Promptable segmentation has recently advanced rapidly due to foundation models such as SAM~\cite{kirillov2023segment}. In parallel, medical foundation segmentation models (e.g., MedSAM~\cite{ma2024medsam} and MedSAM2~\cite{ma2025medsam2}) have demonstrated promising generalization across modalities and anatomies. For volumetric adaptation, SAM-Med3D~\cite{wang2025sam}, Med-SA~\cite{wu2025medical}, and 3DSAM-adapter~\cite{gong20243dsam} extend SAM through 3D architectural or parameter-efficient designs, while SegVol~\cite{du2024segvol} and VISTA3D~\cite{he2025vista3d} provide broader 3D foundation-model alternatives. Despite this progress, transferring promptable segmentation to 3D lesion segmentation remains difficult~\cite{ali2025review}.
Lesions (e.g., liver tumors and enhancing brain tumors) typically occupy a very small portion of a volume and exhibit diverse, irregular morphologies (as shown in Fig.~\ref{fig:1}); both properties amplify ambiguity under sparse prompts and challenge intermediate representations in a transformer image encoder~\cite{xu2021cares,tang2024progressive}.

\begin{figure*}[!t]
\centering
\includegraphics[width=\textwidth]{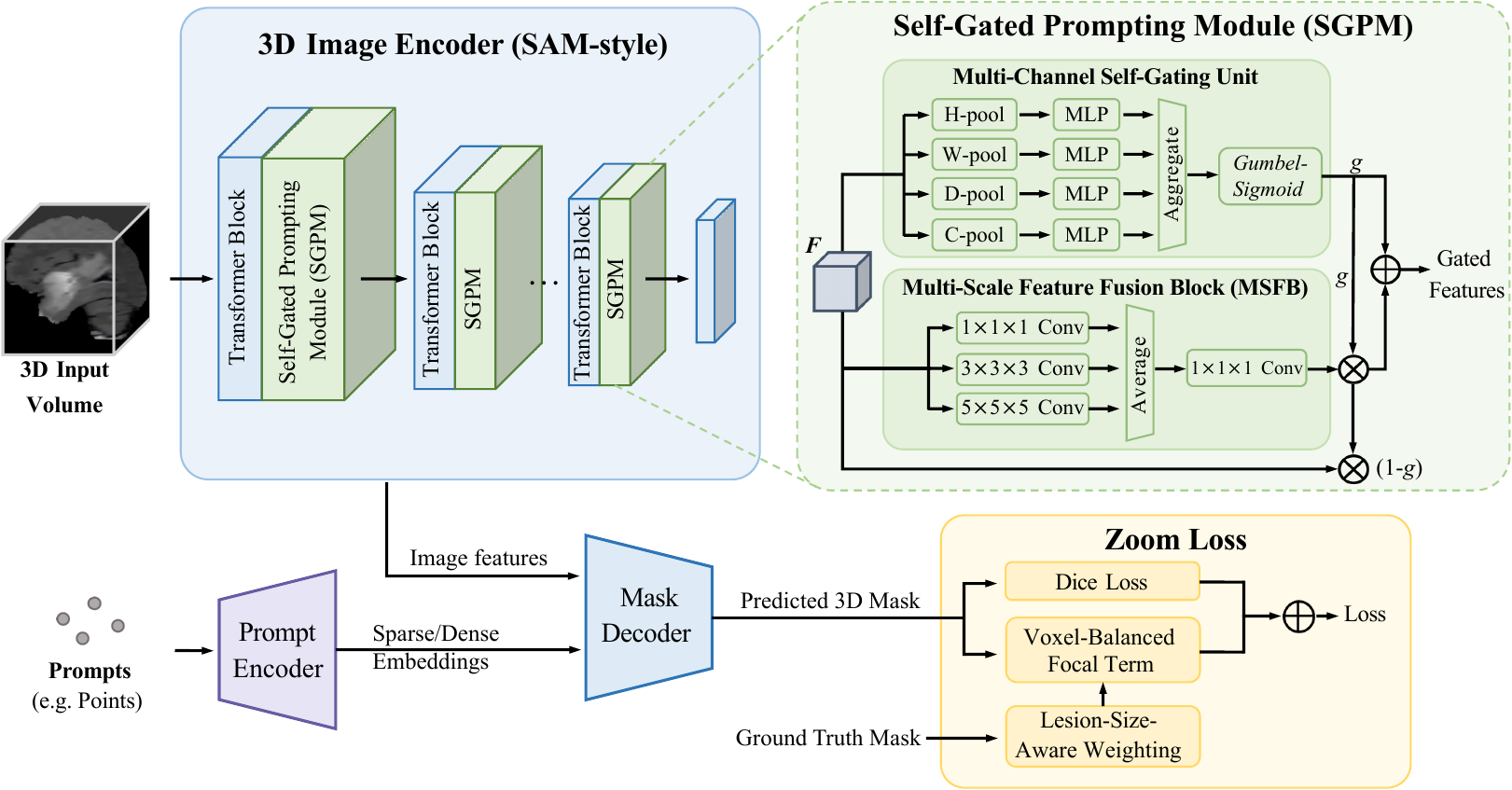}
\caption{SGP-SAM pipeline. SGPM is inserted into the 3D image encoder. A multi-channel gating unit predicts whether to activate MSFB for multi-scale spatial enhancement. Zoom Loss strengthens lesion-focused supervision.}
\label{fig:pipeline}
\end{figure*}

Existing medical SAM adaptations mainly follow two directions. One line focuses on improving 2D medical segmentation by pretraining and then applying per-slice prompting or fine-tuning. Although effective in some settings, slice-wise processing does not fully leverage 3D spatial continuity and may require heavy interaction for volumetric annotation~\cite{ali2025review,huang2025learnable}.
Another line modifies the architecture to process 3D volumes directly (e.g., SAM-Med3D) and explores parameter-efficient transfer (e.g., Med-SA, 3DSAM-adapter)~\cite{li2025stitching}. However, lesion segmentation still suffers from two practical issues: \textbf{(1)} intermediate features may be spatially insufficient for capturing small and complex targets, especially in early-to-mid encoder layers; \textbf{(2)} extreme class imbalance makes training unstable, causing the model to overfit easy background voxels~\cite{ali2025review,xu2021cares}. These observations motivate spatial refinement based on current feature statistics. We therefore use a gate to control the contribution of a multi-scale fusion path. A differentiable relaxation derived from Gumbel-Softmax~\cite{jang2016categorical} makes this binary gate trainable.

To address these issues, we propose \textbf{SGP-SAM}, an adaptation strategy that conditions spatial refinement on current feature statistics. The \textbf{SGPM} is inserted into the 3D image encoder and derives a scalar gate that controls the contribution of a Multi-Scale Feature Fusion Block in a residual path. The scalar gate is broadcast over the feature map and controls a choice at the module level between the enhanced path and the identity path. In addition, we introduce \textbf{Zoom Loss} to strengthen lesion-focused supervision by combining Dice loss~\cite{milletari2016v} with a voxel-balanced focal term~\cite{lin2017focal} and reweighting based on the size of connected components within each volume. 

Our contributions are summarized as follows:
\begin{itemize}
\item We propose \textbf{SGP-SAM}, a feature-gated adaptation framework for transferring 3D SAM-style models to lesion segmentation with improved spatial representation and lesion-focused learning.
\item We design the \textbf{SGPM}, which aggregates dimension-wise feature summaries into an internal scalar gate and controls the contribution of a Multi-Scale Feature Fusion Block in a residual path within the 3D image encoder.

\item We introduce \textbf{Zoom Loss}, which combines Dice loss and voxel-balanced focal supervision with component size weighting within volumes to emphasize smaller lesion components and reduce influence from easy background voxels.
\end{itemize}

\section{Method}
\subsection{Overview}
Figure~\ref{fig:pipeline} illustrates SGP-SAM. We build upon a 3D SAM-style backbone (e.g., SAM-Med3D~\cite{wang2025sam}) consisting of (i) a 3D image encoder, (ii) a prompt encoder, and (iii) a mask decoder.
Given a 3D medical volume $\mathbf{X}\in\mathbb{R}^{H\times W\times D}$, the image encoder outputs latent features
$\mathbf{F}\in\mathbb{R}^{H_d\times W_d\times D_d\times C_d}$.
The prompt encoder converts point prompts into sparse embeddings $\mathbf{S}$ and dense embeddings $\mathbf{D}$, and the mask decoder predicts the final mask $\hat{\mathbf{Y}}\in[0,1]^{H\times W\times D}$.

Our key modification is the \textbf{Self-Gated Prompting Module (SGPM)}, an internal feature-derived controller inserted at the end of selected transformer blocks in the 3D image encoder. SGPM predicts a gate value and conditionally activates a Multi-Scale Feature Fusion Block (MSFB) to enhance spatial representation within the encoder. This end-of-block placement is used for the main results.
Unlike static spatial attention or always-on multi-scale fusion, the gate is predicted from dimension-wise summaries of the current feature map and determines whether MSFB is applied.

\begin{table*}[!t]
\centering
\caption{Results on MSD Liver Tumor and MSD Brain Tumor (enhancing tumor).}
\label{tab:main}
\setlength{\tabcolsep}{10pt}
\small
\renewcommand{\arraystretch}{1.0}
\begin{tabular}{l l l cc}
\toprule
\textbf{Dataset} & \textbf{Method} & \textbf{Init. Model} & \textbf{mIoU} & \textbf{mDice} \\
\midrule
\multirow{4}{*}{MSD Liver Tumor}
& Pretrain & SAM-Med3D & 0.2984 & 0.3993 \\
& Adapter  & SAM-Med3D & 0.4909 & 0.6145 \\
& Fine-tuning & SAM-Med3D & 0.5353 & 0.6667 \\
& \textbf{SGP-SAM (ours)} & SAM-Med3D & \textbf{0.5775} & \textbf{0.7151} \\
\midrule
\multirow{4}{*}{MSD Brain Tumor}
& Pretrain & SAM-Med3D & 0.3576 & 0.4976 \\
& Adapter  & SAM-Med3D & 0.4589 & 0.6029 \\
& Fine-tuning & SAM-Med3D & 0.4437 & 0.5841 \\
& \textbf{SGP-SAM (ours)} & SAM-Med3D & \textbf{0.4653} & \textbf{0.6087} \\
\bottomrule
\end{tabular}
\end{table*}

\subsection{Self-Gated Prompting Module (SGPM)}
SGPM contains: (i) a \textbf{Multi-Channel Self-Gating Unit} that estimates whether intermediate features need additional spatial enhancement, and (ii) an \textbf{MSFB} that performs multi-scale fusion when activated.

\subsubsection{Multi-Channel Self-Gating Unit}
Given $\mathbf{F}\in\mathbb{R}^{H_d\times W_d\times D_d\times C_d}$, we compute dimension-wise summary vectors by mean pooling along complementary axes:
For each $X\in\{H,W,D,C\}$, we retain the $X$-axis and average over the other three axes, producing $\mathbf{f}_X\in\mathbb{R}^{D_X}$, where $(D_H,D_W,D_D,D_C)=(H_d,W_d,D_d,C_d)$.
\begin{equation}
\begin{aligned}
\left[\mathbf{f}_X\right]_j
&=\frac{1}{\prod_{Y\neq X}D_Y}\sum_{\mathbf{i}_{-X}}\mathbf{F}[\mathbf{i}_{-X},j],\\[-1mm]
&\quad j=1,\ldots,D_X,\quad X\in\{H,W,D,C\}.
\end{aligned}
\end{equation}
where $\mathbf{i}_{-X}$ indexes the three averaged axes and $j$ indexes the retained axis.
Each summary is mapped to a scalar key by a two-layer MLP, $k_X=\phi_X(\mathbf{f}_X)$, for $X\in\{H,W,D,C\}$,
where $\phi_\bullet(\cdot)$ outputs a scalar.

We aggregate these keys using learnable weights normalized by softmax:
\begin{equation}
\boldsymbol{\alpha}=\mathrm{softmax}(\mathbf{a})\in\mathbb{R}^4,\quad
s=\alpha_H k_H+\alpha_W k_W+\alpha_D k_D+\alpha_C k_C,
\end{equation}
where $\mathbf{a}$ is a learnable parameter vector and $s$ is the gate logit.
The four summaries provide complementary axis-wise and channel-wise responses, and the softmax weights adapt their contributions to the scalar gate logit.

To make the gate differentiable while encouraging near-binary decisions, we adopt a Gumbel-Sigmoid estimator derived from the Gumbel-Softmax reparameterization:
\begin{equation}
\tilde{g} =
\sigma\left(
\frac{s+\log u-\log(1-u)}{T}
\right),
\quad u\sim\mathcal{U}(0,1).
\end{equation}
where $\sigma(\cdot)$ is the sigmoid and $T$ is a temperature. The gate is used in the output below.
SGPM outputs
\begin{equation}
\mathbf{F}_{\text{out}} = g\cdot \mathrm{MSFB}(\mathbf{F}) + (1-g)\cdot \mathbf{F}.
\label{eq:gated_out}
\end{equation}
Intuitively, if the current features are predicted to be spatially insufficient (gate opens), MSFB enhances them; otherwise, features pass through unchanged.
The scalar gate is broadcast over all entries of $\mathbf{F}$ and controls a module-level choice between the enhanced and identity paths.

\subsubsection{Multi-Scale Feature Fusion Block (MSFB)}
MSFB enhances spatial context by applying parallel 3D convolutions at varying receptive fields to extract multi-scale information~\cite{xu2021cares,hatamizadeh2022swin,tang2024progressive}. We first compress the channels:
we obtain $\mathbf{F}_c=\mathrm{Conv}_{1\times1\times1}(\mathbf{F})\in\mathbb{R}^{H_d\times W_d\times D_d\times C'}$, and apply parallel branches $\mathbf{F}_r=\mathrm{Conv}_{r\times r\times r}(\mathbf{F}_c)$ for $r\in\{1,3,5\}$.
All branches use stride one and same padding, so $\mathbf{F}_1,\mathbf{F}_3,\mathbf{F}_5\in\mathbb{R}^{H_d\times W_d\times D_d\times C'}$. The final $1\times1\times1$ projection restores the channel dimension to $C_d$, yielding $\mathbf{F}_{\mathrm{ms}}\in\mathbb{R}^{H_d\times W_d\times D_d\times C_d}$.
These outputs are fused by averaging, followed by a re-expansion of the channels through a $1\times1\times1$ convolution:
\begin{equation}
\mathbf{F}_{\text{ms}} = \mathrm{ReLU}\left( \mathrm{Conv}_{1\times1\times1}\left( \frac{\mathbf{F}_1 + \mathbf{F}_3 + \mathbf{F}_5}{3} \right) \right),
\label{eq:msfb}
\end{equation}
The first $1\times1\times1$ projection reduces the channel width, while the final projection restores it. The $3\times3\times3$ and $5\times5\times5$ branches provide complementary local and broader spatial contexts. Averaging and re-expanding these responses preserves the feature dimensions, allowing MSFB to enrich representations of small or irregular lesions without changing the encoder interface.
The fusion step provides selective spatial enhancement. In practice, SGPM selectively enhances feature representations by activating MSFB when additional spatial context is needed.
\subsection{Zoom Loss}
The Zoom Loss is designed to address the imbalance in 3D lesion segmentation by assigning larger supervision weights to smaller foreground components. Let $\mathcal{C}=\{C_k\}_{k=1}^{K}$ denote the connected foreground components in a ground-truth volume, $s_k=|C_k|$ their voxel counts, and $\bar{s}=K^{-1}\sum_{k=1}^{K}s_k$ the mean lesion size. For voxel $i$ in component $C_k$, we define
\begin{equation}
 w_i=\left(\frac{\bar{s}+\epsilon}{s_k+\epsilon}\right)^\eta,
 \qquad i\in C_k,
\end{equation}
while background voxels use $w_i=1$. We normalize the weights as $\tilde{w}_i=w_i/(N^{-1}\sum_{j=1}^{N}w_j)$ to preserve the scale of the loss.
Thus, small components receive $w_i>1$, whereas background voxels retain unit weight; normalization preserves the loss scale. The weights are computed only from ground-truth masks during training.
The size-weighted focal term is
\begin{equation}
\scalebox{0.81}{$\displaystyle
\begin{aligned}
 L_{\mathrm{Focal}}=-\frac{1}{N}\sum_{i=1}^{N}\tilde{w}_i\Big[\alpha y_i(1-p_i)^\gamma\log p_i\\
 &+(1-\alpha)(1-y_i)p_i^\gamma\log(1-p_i)\Big],
\end{aligned}
$}
\end{equation}
The supervision also includes the standard soft Dice term $L_{\mathrm{Dice}}$~\cite{milletari2016v}.
The final objective is
\begin{equation}
 L_{\mathrm{Zoom}}=\lambda_D L_{\mathrm{Dice}}+\lambda_F L_{\mathrm{Focal}},
 \qquad \lambda_D+\lambda_F=1.
\end{equation}
Here, $N$ is the number of voxels, $y_i$ and $p_i$ are the ground-truth label and predicted foreground probability, respectively. The parameters $\alpha$ and $\gamma$ control foreground balancing and hard-example focusing, while $\eta$ controls the strength of lesion-size reweighting. We use $\alpha=0.25$, $\gamma=2$, $\lambda_D=\lambda_F=0.5$, $\eta=0.5$, and $\epsilon=10^{-6}$.
During training, component sizes and reweighting factors are computed from each ground-truth mask; no loss weights are required at inference.


\section{Experiment}

\subsection{Dataset}
\textbf{MSD Liver Tumor}~\cite{simpson2019large,bilic2023liver} is one of the datasets used in the Medical Segmentation Decathlon challenge. Its training set consists of 3D CT images of the liver from 131 patients and corresponding annotations of liver tumors. We randomly split the dataset into 80\% training and 20\% testing cases. All SAM-Med3D-based methods use the same preprocessing protocol.
\\
\textbf{MSD Brain Tumor}~\cite{simpson2019large,menze2014multimodal}. Its training set consists of 484 brain MRI images, featuring three types of labels: edema, enhancing tumor, and non-enhancing tumor. We chose the enhancing tumor, which has the most complex structures and presents the greatest segmentation challenge, as our target for experimentation. We randomly divided the dataset into 80\% training and 20\% testing cases. We applied the same data preprocessing method to this dataset.

\subsection{Implementation Details}
All trainable settings were optimized with AdamW using a learning rate of 8e-4 for 40 epochs, a batch size of 16, and a weight decay of 1e-5; the random seed was fixed to 42. We monitored the loss to ensure training convergence. All experiments were conducted on an Nvidia 4090. All transfer settings in Table~\ref{tab:main} use SAM-Med3D as the base model: ``Pretrain'' denotes direct use of the pretrained model, ``Adapter'' denotes parameter-efficient adapter tuning, and ``Fine-tuning'' denotes full model adaptation. During training, each prompt-based sample used one positive point sampled inside the lesion and one negative point sampled from the background. At test time, the same fixed prompt set was used for all SAM-based methods. For SGP-SAM, the pretrained SAM-Med3D weights initialized the model, and the SGPM and mask decoder were optimized while the pretrained prompt encoder remained frozen. We retain SAM-Med3D as a common backbone to isolate transfer strategy; all variants share the same split, preprocessing, and prompt protocol.
We report mIoU and mDice on the held-out 20\% test split.

\subsection{Comparisons with Related Methods}

The comparative experiments show consistent improvements over the selected transfer baselines~\cite{wang2025sam}. We compare SGP-SAM with SAM-Med3D pretraining, adapter tuning, and full fine-tuning. After training with Zoom Loss on the MSD Liver Tumor dataset, SGP-SAM reaches 0.5775 mIoU and 0.7151 mDice. Compared with SAM-Med3D fine-tuning, it improves mIoU and mDice by 7.9\% and 7.3\%, respectively. On MSD Brain Tumor, SGP-SAM improves over fine-tuning by 4.9\% mIoU and 4.2\% mDice.
On Liver, it exceeds fine-tuning by 0.0422 mIoU and 0.0484 mDice, and adapter tuning by 0.0866 and 0.1006. On Brain, it remains best in both metrics, with mDice gains of 0.0058 and 0.0246 over adapter tuning and fine-tuning.

\begin{figure}[t]
\centering
\includegraphics[width=0.8\linewidth]{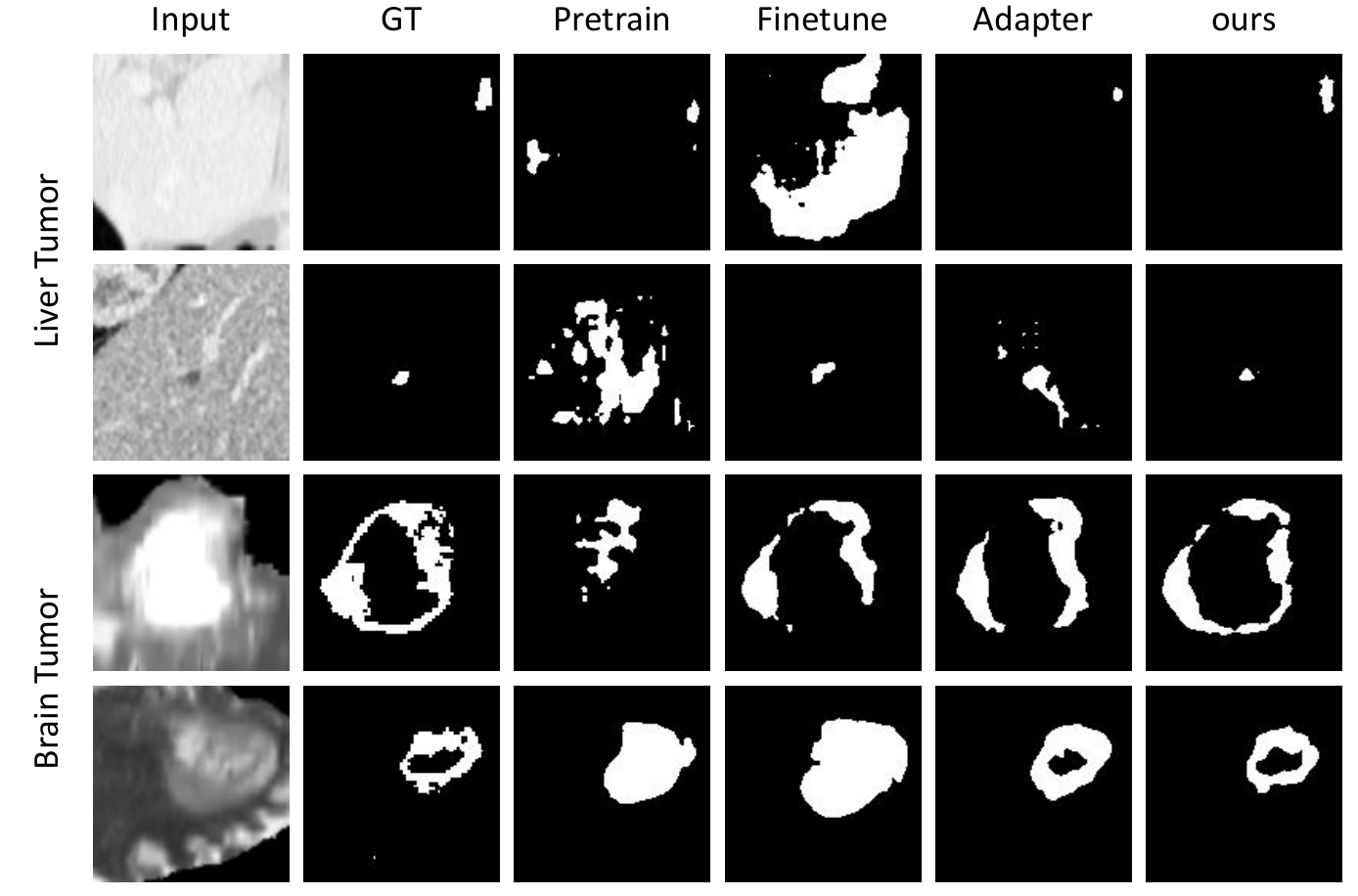}
\caption{The visualization results demonstrate that our SGP-SAM method achieves better segmentation performance for both liver tumors with small volumes and brain tumors with complex shapes and ambiguous boundaries.}
\label{fig:3} 
\end{figure}

The visualization results further demonstrate the effectiveness of our method. As shown in Fig.~\ref{fig:3}, for liver tumors with small proportions and blurry boundaries, benefiting from Zoom Loss, the model can pay more attention to smaller lesion areas and achieve more accurate segmentation. For brain tumors with complex structures and ambiguous boundaries, the multi-scale enhancement provided by SGPM is consistent with the improved structural delineation. By extracting features at multiple scales using compressed information from various dimensions, the module enriches feature representations for these lesion areas.

\begin{figure}[t!]
\centering
\includegraphics[width=\linewidth]{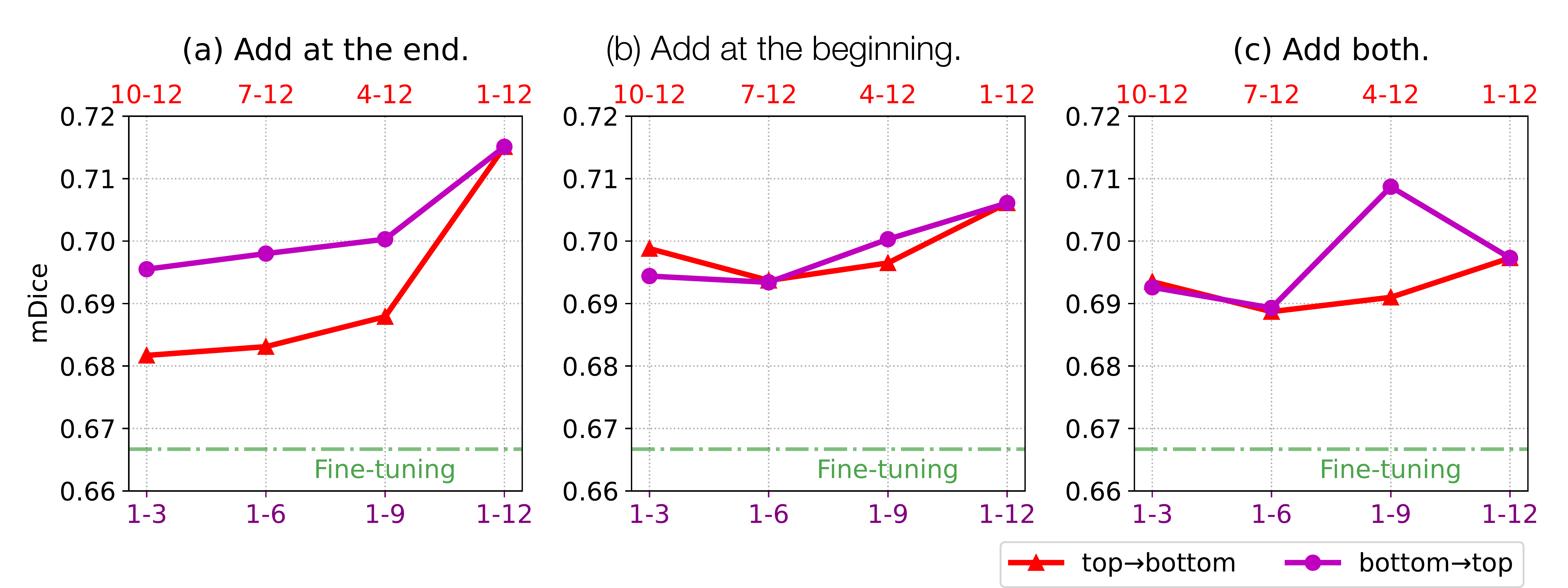}
\caption{Ablation study on SGPM depth and position.  (a) SGPM is added at the end of the block. (b) SGPM is added at the beginning of the block. (c) SGPM is added on both sides of the block. ``$i-j$'' denotes that the SGPM is embedded from the $i$-th layer to the $j$-th layer of the image encoder.
 Note: the 1st layer of the image encoder is the bottommost layer.}
\label{fig:4} 
\end{figure}

\begin{table}[t]
\centering
\caption{Ablation study of MSFB, Gate, and Zoom Loss. A dash denotes a disabled module.}
\setlength{\tabcolsep}{2pt}
\begin{tabular*}{\columnwidth}{@{\extracolsep{\fill}}ccc|cc@{}}
\hline
\textbf{MSFB} & \textbf{Gate} & \textbf{Zoom Loss} & \textbf{mIoU} & \textbf{mDice} \\ \hline
-- & -- & -- & 0.5353 & 0.6667 \\
\checkmark & -- & -- & 0.5427 & 0.6724 \\
-- & -- & \checkmark & 0.5546 & 0.6783 \\
\checkmark & \checkmark & -- & 0.5585 & 0.6931 \\
\checkmark & \checkmark & \checkmark & 0.5775 & 0.7151 \\ \hline
\end{tabular*}

\label{table:ablation}
\end{table}

\subsection{Ablation Studies}
Ablation results on the MSD Liver Tumor dataset show the separate contributions of conditional spatial enhancement and lesion-focused supervision. The row with all modules disabled corresponds to the SAM-Med3D full fine-tuning baseline. As shown in Tab.~\ref{table:ablation}, always-on MSFB reaches 0.5427 mIoU and 0.6724 mDice, whereas gated SGPM reaches 0.5585 and 0.6931. The gated design therefore improves over uniform MSFB activation by 0.0158 mIoU and 0.0207 mDice, indicating that the gain is not explained only by adding the multi-scale fusion block. Zoom Loss alone reaches 0.5546 mIoU and 0.6783 mDice. Adding SGPM to Zoom Loss further improves mIoU and mDice by 0.0229 and 0.0368, respectively, reaching 0.5775 mIoU and 0.7151 mDice. These results indicate complementary benefits from conditional spatial enhancement and lesion-focused supervision.
The shared backbone and optimization protocol make the always-on versus gated contrast isolate input-dependent gating.

The ablation study on SGPM depth and position (Fig.~\ref{fig:4}) reveals that (1) model performance improves with the SGPM module; (2) adding SGPM from bottom to top layers is more effective, enhancing spatial representation in early layers; (3) placing SGPM at the end of the block yields the best performance. Hence, adding SGPM at the end of each block provides the most effective configuration among the evaluated settings.

\section{Conclusion}
{SGP-SAM addresses two recurring difficulties in transferring promptable 3D foundation models to lesion segmentation: insufficient spatial context for small or irregular targets and severe foreground-background imbalance. SGPM summarizes intermediate features along spatial and channel axes, learns a feature-derived gate, and selects between an MSFB-enhanced path and the identity path while preserving the encoder interface. Zoom Loss complements this refinement with lesion-size-aware focal and Dice supervision. On both MSD tasks, SGP-SAM improves over SAM-Med3D pretraining, adapter tuning, and full fine-tuning; it reaches 0.7151 mDice on Liver and a 7.3\% relative gain over fine-tuning. Ablations show that gated enhancement outperforms always-on fusion, while SGPM and Zoom Loss provide complementary gains.\par}
\bibliographystyle{IEEEbib}
\bibliography{ref}

@article{kirillov2023segment,
  title={Segment anything},
  author={Kirillov, Alexander and Mintun, Eric and Ravi, Nikhila and Mao, Hanzi and Rolland, Chloe and Gustafson, Laura and Xiao, Tete and Whitehead, Spencer and Berg, Alexander C and Lo, Wan-Yen and others},
  journal={arXiv:2304.02643},
  year={2023}
}

@inproceedings{lin2017focal,
  title={Focal loss for dense object detection},
  author={Lin, Tsung-Yi and Goyal, Priya and Girshick, Ross and He, Kaiming and Doll{\'a}r, Piotr},
  booktitle={Proceedings of the IEEE international conference on computer vision},
  pages={2980--2988},
  year={2017}
}

@article{jang2016categorical,
  title={Categorical reparameterization with gumbel-softmax},
  author={Jang, Eric and Gu, Shixiang and Poole, Ben},
  journal={arXiv:1611.01144},
  year={2016}
}

@article{menze2014multimodal,
  title={The multimodal brain tumor image segmentation benchmark (BRATS)},
  author={Menze, Bjoern H and Jakab, Andras and Bauer, Stefan and Kalpathy-Cramer, Jayashree and Farahani, Keyvan and Kirby, Justin and Burren, Yuliya and Porz, Nicole and Slotboom, Johannes and Wiest, Roland and others},
  journal={IEEE transactions on medical imaging},
  volume={34},
  number={10},
  pages={1993--2024},
  year={2014},
  publisher={IEEE}
}

@article{bilic2023liver,
  title={The liver tumor segmentation benchmark (lits)},
  author={Bilic, Patrick and Christ, Patrick and Li, Hongwei Bran and Vorontsov, Eugene and Ben-Cohen, Avi and Kaissis, Georgios and Szeskin, Adi and Jacobs, Colin and Mamani, Gabriel Efrain Humpire and Chartrand, Gabriel and others},
  journal={Medical Image Analysis},
  volume={84},
  pages={102680},
  year={2023},
  publisher={Elsevier}
}

@article{simpson2019large,
  title={A large annotated medical image dataset for the development and evaluation of segmentation algorithms},
  author={Simpson, Amber L and Antonelli, Michela and Bakas, Spyridon and Bilello, Michel and Farahani, Keyvan and Van Ginneken, Bram and Kopp-Schneider, Annette and Landman, Bennett A and Litjens, Geert and Menze, Bjoern and others},
  journal={arXiv:1902.09063},
  year={2019}
}

@inproceedings{milletari2016v,
  title={V-net: Fully convolutional neural networks for volumetric medical image segmentation},
  author={Milletari, Fausto and Navab, Nassir and Ahmadi, Seyed-Ahmad},
  booktitle={2016 fourth international conference on 3D vision (3DV)},
  pages={565--571},
  year={2016},
  organization={Ieee}
}

@article{hatamizadeh2022swin,
  title={Swin UNETR: Swin Transformers for Semantic Segmentation of Brain Tumors in MRI Images},
  author={Hatamizadeh, Ali and Nath, Vishwesh and Tang, Yucheng and Yang, Dong and Roth, Holger and Xu, Daguang},
  journal={arXiv:2201.01266},
  year={2022}
}

@article{ma2024medsam,
  title={Segment anything in medical images},
  author={Ma, Jun and He, Yuting and Li, Feifei and Han, Lin and You, Chenyu and Wang, Bo},
  journal={Nature Communications},
  volume={15},
  number={1},
  pages={654},
  year={2024},
  publisher={Nature Publishing Group}
}

@inproceedings{du2024segvol,
  title={SegVol: Universal and Interactive Volumetric Medical Image Segmentation},
  author={Du, Yuxiang and Ye, Jin and Deng, Zhongying and others},
  booktitle={Advances in Neural Information Processing Systems (NeurIPS)},
  year={2024}
}

@InProceedings{he2025vista3d,
  author = {He, Yufan and Guo, Pengfei and Tang, Yucheng and Myronenko, Andriy and Nath, Vishwesh and Xu, Ziyue and Yang, Dong and Zhao, Can and Simon, Benjamin and Belue, Mason and Harmon, Stephanie and Turkbey, Baris and Xu, Daguang and Li, Wenqi},
  title = {VISTA3D: A Unified Segmentation Foundation Model For 3D Medical Imaging},
  booktitle = {Proceedings of the IEEE/CVF Conference on Computer Vision and Pattern Recognition (CVPR)},
  month = {June},
  year = {2025},
  pages = {20863--20873}
}

@article{wang2025sam,
  title={SAM-Med3D: a vision foundation model for general-purpose segmentation on volumetric medical images},
  author={Wang, Haoyu and Guo, Sizheng and Ye, Jin and Deng, Zhongying and Cheng, Junlong and Li, Tianbin and Chen, Jianpin and Su, Yanzhou and Huang, Ziyan and Shen, Yiqing and others},
  journal={IEEE Transactions on Neural Networks and Learning Systems},
  year={2025},
  publisher={IEEE}
}

@article{gong20243dsam,
  title={3dsam-adapter: Holistic adaptation of sam from 2d to 3d for promptable tumor segmentation},
  author={Gong, Shizhan and Zhong, Yuan and Ma, Wenao and Li, Jinpeng and Wang, Zhao and Zhang, Jingyang and Heng, Pheng-Ann and Dou, Qi},
  journal={Medical Image Analysis},
  volume={98},
  pages={103324},
  year={2024},
  publisher={Elsevier}
}

@article{ma2025medsam2,
  title={Medsam2: Segment anything in 3d medical images and videos},
  author={Ma, Jun and Yang, Zongxin and Kim, Sumin and Chen, Bihui and Baharoon, Mohammed and Fallahpour, Adibvafa and Asakereh, Reza and Lyu, Hongwei and Wang, Bo},
  journal={arXiv preprint arXiv:2504.03600},
  year={2025}
}

@article{ali2025review,
  title={A review of the segment anything model (sam) for medical image analysis: Accomplishments and perspectives},
  author={Ali, Mudassar and Wu, Tong and Hu, Haoji and Luo, Qiong and Xu, Dong and Zheng, Weizeng and Jin, Neng and Yang, Chen and Yao, Jincao},
  journal={Computerized Medical Imaging and Graphics},
  volume={119},
  pages={102473},
  year={2025},
  publisher={Elsevier}
}

@article{wu2025medical,
  title={Medical sam adapter: Adapting segment anything model for medical image segmentation},
  author={Wu, Junde and Wang, Ziyue and Hong, Mingxuan and Ji, Wei and Fu, Huazhu and Xu, Yanwu and Xu, Min and Jin, Yueming},
  journal={Medical image analysis},
  volume={102},
  pages={103547},
  year={2025},
  publisher={Elsevier}
}

@article{li2025stitching,
  title={Stitching, Fine-Tuning, and Re-Training: A SAM-Enabled Framework for Semi-Supervised 3D Medical Image Segmentation},
  author={Li, Shumeng and Qi, Lei and Yu, Qian and Huo, Jing and Shi, Yinghuan and Gao, Yang},
  journal={IEEE Transactions on Medical Imaging},
  volume={44},
  number={10},
  pages={3909--3923},
  year={2025},
  publisher={IEEE}
}

@article{huang2025learnable,
  title={Learnable prompting sam-induced knowledge distillation for semi-supervised medical image segmentation},
  author={Huang, Kaiwen and Zhou, Tao and Fu, Huazhu and Zhang, Yizhe and Zhou, Yi and Gong, Chen and Liang, Dong},
  journal={IEEE Transactions on Medical Imaging},
  volume={44},
  number={5},
  pages={2295--2306},
  year={2025},
  publisher={IEEE}
}

@article{tang2024progressive,
  title={Progressive deep snake for instance boundary extraction in medical images},
  author={Tang, Zixuan and Chen, Bin and Zeng, An and Liu, Mengyuan and Zhao, Shen},
  journal={Expert Systems with Applications},
  volume={249},
  pages={123590},
  year={2024},
  publisher={Elsevier}
}

@article{xu2021cares,
  title={CARes-UNet: Content-aware residual UNet for lesion segmentation of COVID-19 from chest CT images},
  author={Xu, Xinhua and Wen, Yuhang and Zhao, Lu and Zhang, Yi and Zhao, Youjun and Tang, Zixuan and Yang, Ziduo and Chen, Calvin Yu-Chian},
  journal={Medical physics},
  volume={48},
  number={11},
  pages={7127--7140},
  year={2021},
  publisher={Wiley Online Library}
}

\end{document}